# Enhancing Medical Support in the Arabic Language Through Personalized ChatGPT Assistance

Mohamed Issa[1,2] and Ahmed Abdelwahed[1]

[1]Computer and Systems Department, Faculty of Engineering, Zagazig University, Egypt
[2]Faculty of Computer Science and Information Technology, Egypt-Japan University for Science and Technology, Egypt

*Abstract*— **This Paper discusses the growing popularity of online medical diagnosis as an alternative to traditional doctor visits. It highlights the limitations of existing tools and emphasizes the advantages of using ChatGPT, which provides real-time, personalized medical diagnosis at no cost. The paragraph summarizes a research study that evaluated the performance of ChatGPT in Arabic medical diagnosis. The study involved compiling a dataset of disease information and generating multiple messages for each disease using different prompting techniques. ChatGPT's performance was assessed by measuring the similarity between its responses and the actual diseases. The results showed promising performance, with average scores of around 76% for similarity measures. Various prompting techniques were used, and chain prompting demonstrated a relative advantage. The study also recorded an average response time of 6.12 seconds for the ChatGPT API, which is considered acceptable but has room for improvement. While ChatGPT cannot replace human doctors entirely, the findings suggest its potential in emergency cases and addressing general medical inquiries. Overall, the study highlights ChatGPT's viability as a valuable tool in the medical field.**

## I. Introduction

The internet is increasingly becoming a popular resource for people to research their health concerns. In the UK, for example, the NHS Choices online patient portal for national health information reports over 15 million visits per month [1]. In the United States, more than a third of adults regularly use the internet to self-diagnose their ailments, including both non-urgent and urgent symptoms such as chest pain. However, relying on search engines like Google, Bing, or Yahoo to self-diagnose can lead to confusion and sometimes unsubstantiated information. Moreover, people with urgent symptoms may not be directed to seek emergent care through internet searches. While there are numerous online resources available to learn about specific conditions, caution should be exercised when using the internet for self-diagnosis [2, 3]. In recent times, there has been a proliferation of more sophisticated programs called symptom checkers that aim to more effectively provide potential diagnoses for patients and direct them to the appropriate care setting [4].

Visiting a doctor in a clinic or hospital is one of the traditional methods for medical consultations. The main advantages of this way include personalized diagnosis and high diagnostic accuracy due to the doctor's ability to examine the patient physically and consider their medical history. In case of emergency situations, this method may be the best option, as it allows for immediate medical attention and treatment. This method has its limitations such as being time-consuming and costly, especially for patients who live far away from medical facilities. Additionally, patients may be required to take time off from work or school to attend their appointments, scheduling an appointment and traveling to the location.

The requirement of visiting a doctor in person is a sub-optimal solution, as visiting a doctor involves reserving part of your day for the appointment. And often doctors are not always present so you will have to try and find a time slot where you and your doctor are available. This can be a hard problem especially if you both have busy schedules. With the recent spread of viral diseases, a new downside of visiting doctors in person emerged as one can catch a disease that can potentially threaten their life. All these shortcomings motivated the development of tele-medicine websites connecting patients with a doctor from the comfort of their homes. With less cost and more availability telemedicine websites are helping many people get their medical diagnosis without the hassle of a doctor's visit. Although tele-medicine is an improvement to physically visiting a doctor it's still far from convenient as you are still required to make an appointment and you a have a relatively pricey bill to fulfill. Another limitation for this approach is the lack of flexibility in appointment times. As well as its costly nature.

To overcome the main limitations of visiting a doctor and tele-medicine, symptom checkers were developed. Symptom checkers use computerized algorithms to ask users a series of questions about their symptoms or require them to input details about their symptoms themselves. The algorithms vary and may use branching logic, Bayesian inference, or other methods. Private companies and organizations, including the National Health Service, the American Academy of Pediatrics, and the Mayo Clinic, have launched their own symptom checkers. One example, iTriage, reports 50 million uses each year. These tools are typically accessed through websites, but some are also available as apps for smartphones or tablets [5].

Symptom checkers serve two main functions: to facilitate self-diagnosis and to assist with triage. The self-diagnosis function provides a list of potential diagnoses, usually ranked by likelihood. This function is typically framed as helping educate patients on the range of diagnoses that might fit their symptoms. The triage function informs patients whether they should seek care at all and, if so, where (such as emergency department or general practitioner's clinic) and with what urgency (such as immediately or within a few days). Symptom checkers typically cater to three healthcare needs of a patient. Firstly, they provide information, wherein a patient seeks to learn more about the symptoms or conditions they may have. Secondly, patients may want to know whether their symptoms require treatment or further investigation, and symptom checkers provide medical triage by directing patients to the most suitable location within an appropriate time frame. The appropriate action depends on the nature and urgency of the symptoms or their underlying cause, which might require further investigation. Finally, patients may want to understand the conditions that might be responsible for their symptoms. This corresponds to diagnosis or "differential diagnosis" and is typically performed by an experienced medical practitioner. While symptom checkers can be helpful for providing information and triage, they should not be used as a substitute for a proper medical diagnosis from a qualified healthcare professional.

The potential benefits of symptom checkers are several. They can encourage patients with life-threatening problems such as stroke or heart attack to seek emergency care immediately. For patients with a non-emergent problem that does not require a medical visit, these programs can provide reassurance and recommend they stay home, avoiding unnecessary visits to healthcare providers. By providing patients with detailed information about their symptoms, symptom checkers can also increase patient engagement and empowerment in their own health management. However, it is important to note that symptom checkers should not replace the advice and diagnosis of a qualified healthcare professional, and patients should always seek medical attention when they have concerning symptoms [6].

Reducing the number of visits for acute respiratory illness such as viral upper respiratory tract infection can save patients' time and money, reduce the over prescription of antibiotics, and decrease the demand on primary care providers. It is worth noting that approximately a quarter of visits for these illnesses result in no intervention beyond over-the-counter treatment, and over half of patients receive unnecessary antibiotics. This is a significant problem, especially considering that the workload for general practitioners in the UK increased by 62% from 1995 to 2008. By using symptom checkers to triage and provide self-diagnosis, patients with nonemergent illnesses can receive appropriate advice and treatment, reducing the burden on healthcare providers [7].

There are several key concerns associated with using symptom checkers. One concern is that if patients with life-threatening problems are misdiagnosed and not directed to seek care, their health could worsen, leading to increased morbidity and mortality. Conversely, if patients with minor illnesses are advised to seek care, particularly in an emergency department, these programs could increase unnecessary visits, resulting in increased time and costs for patients and society. Additionally, symptom checkers may not be suitable for all patients, particularly those who have difficulty understanding the questions or inputting their symptoms. It is important to note that symptom checkers should not replace the advice and diagnosis of a qualified healthcare professional, and patients should always seek medical attention when they have concerning symptoms.

The main limitations of checkers are concluded as follows:

• Only useful for mild and non-serious medical cases.
• Suffers low diagnostic accuracy.
• Limits users to a set of static options to describe their condition.
• Not personalized and does not consider the patient's medical history.
• Some of them are biased as they are supported by certain pharmaceutical companies.

In the field of artificial intelligence, the development of conversational AI has been one of the most exciting breakthroughs in recent years. Generative Pretrained Transformers (GPT) are systems that can understand and generate complex concepts in long strings of text. ChatGPT, an NLP model developed by OpenAI, allows for real-time conversations with an AI chatbot that closely resembles human interaction. The model is based on the GPT architecture, which uses unsupervised learning to generate writing that is similar to human language. ChatGPT collects data from various sources, feeds it into a transformer model, maps the connections between the different pieces of data, and makes informed predictions about what text should be used in particular situations [8, 9].

As technology continues to evolve, ChatGPT and similar technologies can be trained on organizational data to transform industries. These technologies serve as excellent starting points for developing software and content, managing knowledge, improving consumer interactions, and enhancing employee experiences. In the future, event planners, tutors, and virtual personal assistants are among the potential developments that could benefit from these technologies [10, 11]. ChatGPT's success can be attributed to its ability to provide highly accurate and relevant responses to users' queries, which has made it a go-to tool for many people. The chat bot's natural language processing capabilities enable it to understand the nuances of language and provide responses that are not only accurate but also contextually relevant.

While many aspects of healthcare require a connection with patients, it is not always necessary for optimal treatment. By improving adherence to treatment regimens and providing more practical and accessible care, ChatGPT can enhance the care provided by human healthcare providers and improve patient outcomes [12, 13]. To be used in healthcare, ChatGPT's prediction capabilities would need to be controlled. Transformer models can detect patterns in the training data and apply that knowledge during inference, potentially leading to hallucinated predictions in medical summaries. However, ChatGPT can be a valuable resource for patients who live in underserved or rural areas and have difficulty physically meeting with certified diabetes educators or other healthcare experts. Such patients could turn to ChatGPT as a reliable source of help and information, even if they cannot physically visit a healthcare center [14-16]. ChatGPT can offer an improved healthcare solution that facilitates communication between medical care providers and patients [17-20].

## II. CHATGPT

Large Language Models (LLMs) have seen a recent surge after the development of the transformer model architecture [21]. Many advancements have been made since then, but the most potent was the Reinforcement Learning from Human Feedback or (RLHF) breakthrough as this was an effective way to steer the model output to be more suitable for human preferences [22]. This allowed the creation of a model that is able to chat in meaningful conversations giving accurate and articulate responses. ChatGPT was received with massive popularity, denoting the big value these types of models can offer to the general public. Many use cases have emerged from ChatGPT ranging from simple questions to assisting professionals [23]. One of the most helpful ways ChatGPT can be used is the medical field where it can help someone feel better or even save their life [19].

ChatGPT has already demonstrated its some reliability in medical tasks, as evidenced by its successful completion of the United States Medical Licensing Exam [24]. Furthermore, it has surpassed human physicians in the quality of its responses [25]. The potential applications of ChatGPT in medicine are numerous and promising. In particular, its utilization as a medical assistant holds great potential in the upcoming years. At present, modifications to ChatGPT can only be achieved through prompt engineering. There are various approaches to providing the model with sufficient information and seeking specific responses. The question at hand is: what is the correct method to accomplish this?

The popularity of ChatGPT is expected to continue to grow as more businesses and individuals recognize its potential. The chat bot is already being used by companies to improve their customer service experience and by content creators to generate high-quality content quickly and efficiently. Hence, ChatGPT is an innovative tool that represents the cutting edge of conversational AI. Its advanced deep learning techniques and natural language processing capabilities make it a highly versatile and effective tool that has already gained massive popularity among users. As AI technology continues to evolve, the potential of chat bots like ChatGPT is limitless, and we can expect to see more exciting developments in this field in the future.

The use cases of ChatGPT have been various from presentations to creative writing and coding. However, its usage in the healthcare sector raises ethical and legal concerns, particularly if the general public or medical professionals rely on ChatGPT for treatment decisions. ChatGPT can understand the general context of a query or conversation and provide detailed responses that are relevant to the subject. In the healthcare sector, ChatGPT can be leveraged to achieve multiple goals, such as improving patient experiences, assisting medical personnel in optimizing healthcare procedures, and uncovering valuable information.

ChatGPT can understand and respond to different conversational inputs, such as queries, claims, and directives in the healthcare domain. Its ability to converse with patients naturally and human-like is a significant advantage for chatbots, customer service agents, and digital assistants in the healthcare sector. ChatGPT leverages Machine Learning (ML) algorithms and natural language processing (NLP) approaches to enhance its conversational capabilities. The model can analyze and predict word sequences by utilizing word embedding and Transformers. Continuous feeding of data into chatbot neural networks is critical for effective AI, and ML plays a crucial role in achieving that [26, 27]. By analyzing various written materials, including textual and spoken language, ChatGPT can provide accurate responses to customer inquiries and frequently asked questions [28, 29]. ChatGPT is most effective when used as a chatbot to offer customer support, answer questions, and perform other duties in various businesses [30].

ChatGPT can aid in analyzing customer data and segmenting customers based on their preferences, needs, and behaviors, enabling the marketing team to conduct more successful targeted marketing campaigns. With its advanced machine-learning algorithms, ChatGPT can overcome language barriers, significantly enhancing consumer experiences and expanding a company's worldwide reach. It can also help companies strengthen their online presence by being accessible around the clock on websites and social media platforms, enabling quick responses to consumers' frequently asked questions [24, 31].

## III. MESSAGE GENERATION TECHNIQUES

Real-life disease data were collected for 1144 diseases. Data was collected from the Mayo Clinic Arabic website then stored in a local database. Mayo Clinic ensures the precision of its medical information through thousands of medical experts and scientists. From which about 100 serve as medical editors working to ensure the correctness and reliability of the information provided. An automated script was run to crawl each disease's name and all associated symptoms. Symptoms are used to formulate messages for ChatGPT which predicts the likely disease. The diagnosis from the model is compared to the ground truth collected from the website.

Since the release of ChatGPT, many practitioners shared effective prompt structures that proved useful in guiding the model to generate accurate results [32]. Simply writing keywords to get a good response may be useful when working with search engines, but this is far from the optimal practice when it comes to large language models. Utilizing the attention mechanism LLMs can effectively attend to the whole sentence enabling the model to remember every query word while generating the response [33]. Having the ability to recall the whole context allows for more complex and long prompts than just keywords. These prompting methods were developed either through experimentation or through an understanding of the underlying model. Some techniques that proved effective are:

*More information equals better responses:* More context almost always leads to a better response as the model has more knowledge to make better decisions formulating a more useful answer [34]. ChatGPT has a large context size of 4K tokens, and it can remember all of them while generating the answer. Not utilizing the large context size is sacrificing a better and more tailored answer. In this work,15 different messages were created, each with increasingly more context. In our case more context means more symptoms.

*Model acts in a certain role:* Large Language Models (LLMs) are trained on massive amounts of text data written by different people with diverse backgrounds. Asking the LLM to act as a certain occupation, places it in the right mindset before answering. Ultimately, LLMs are trained to predict the next word in a document. Models are trained on data from many sources, each from a different field. Hinting which field, we want the model to think about comes with huge benefits as it places the model in the right "frame of mind" [9]. For example, starting the message with "You are an experienced doctor giving accurate medical advice.". effectively makes the model recall data during its training when there was an experienced doctor giving a quality answer.

*Chain prompting the model:* Giving the model an example (prompt, response) before providing your prompt helps to shape the model response [35]. Long prompts are created in this technique containing one or multiple prompt and response examples before giving the model the intended prompt to answer. Chain prompting proved to give a more accurate response with rare sparsity. One might think of this method as imitating the structure of a unique Q and A website. The model understands that there is a high probability of the text following the same structure.

An example of implementing this method is:

*"Prompt: What are the three main colors?*

*Response: [Red, Green, Blue]*

*Prompt: List the names of rivers in Africa.*

*Response: [The Nile, The Congo River, The Niger river]*

*Prompt: What are the old-world continents?"*

The answer is very likely to be in the same form as the former example. In fact, the ChatGPT answer to this prompt was: *"[Africa, Asia, and Europe]"*. Therefore, this method enables Steering the model in the aspired format. Prompt engineering is a field in the infancy stage. Many prompting methods were discovered to be effective in general, but the best one is left to be decided by experimentation. To measure the performance of ChatGPT across these methods, multiple messages were generated for each disease. Each message is an implementation of a prompting method. For a disease we formulate eight different messages representing every prompting method.

## IV. EXPERIMENTAL RESULTS AND DISCUSSION

To collect disease data a python script was developed looping through all disease pages on the Mayo Clinic Arabic website and storing each disease name and symptoms locally. First a session is created using a Chrome browser user agent [36]. Starting from the page containing links of disease lists based on the initial letter of the disease name, the link for each letter list is collected and stored in an array. Each link contains a page of all disease URLs starting with a particular letter. The next step was to collect all disease links and store them in a list. Loading every disease page, we can now - with the use of the BeautifulSoup Python library - locate the HTML tags containing the disease title and its symptoms. We stored each disease data as pickled python dictionary.

Messages were generated from each disease with different methods. The first group of methods was testing the model performance when providing different numbers of symptoms, so we increased the number of symptoms in every message each time by one. Starting from one symptom to six symptoms per message also messages were created containing all symptoms of the disease which were 18 at max. The second group contained three methods. First, a method focused on providing the model with the largest possible context size. The second method was asking the model to play a certain role, particularly an expert doctor. While the third was providing a chain of prompt-response examples before inserting the prompt we want the model to answer.

TABLE I. SHOWS MESSAGES USING DIFFERENT PROMPT ENGINEERING TECHNIQUES ALONG WITH THE CORRESPONDING CHATGPT RESPONSE.

| Method | Message | Response |
|---|---|---|
| More Information (3 symptoms) | من فضلك ساعدني اعاني من الاعراض التالية التعرّق والقشعريرة والارتجاف والصداع اخبرني بالمرض الذي اعاني منه | بناءً على الأعراض التي ذكرتها يمكن ان تكون تعاني من التهاب الحمى الروماتيزمية |
| More Information (4 symptoms) | اشعر بالمرض واعاني من تشوش الرؤية والدوار وخفقان القلب والضعف اخبرني بالمرض الذي اعاني منه | التهاب الأذن الوسطى |
| More Information (5 symptoms) | اشعر بالمرض واعاني من الحمى واسهال والإرهاق وآلام في العضلات وسعال اخبرني بالمرض الذي اعاني منه | التهاب الشعب الهوائية الحاد |
| More Information (6 symptoms) | من فضلك ساعدني اعاني من الاعراض التالية الألم، وخاصة عند تحميل الوزن على القدم المصابة والإيلام عند لمس الكاحل والتورم والكدمات وتقييد نطاق الحركة وعدم ثبات الكاحل اخبرني بالمرض الذي اعاني منه | بناءً على الأعراض التي وصفتها، يمكن أن يكون لديك التهاب اللفافة الأخمصية (Plantar Fasciitis). |
| More Information (all symptoms) | مؤخرا وجدت نفسي تعاني من الميل إلى الإمساك بالمواد المقروءة على بُعد لجعل الحروف أكثر وضوحاً وعدم وضوح الرؤية عند المسافة الطبيعية للقراءة وإجهاد العين أو الصداع بعد القراءة أو تنفيذ الأعمال عن قرب اخبرني بالمرض الذي اعاني منه | بناءً على الأعراض التي وصفتها، يمكن أن يكون لديك ما يُعرف بـ "قصر النظر" أو "الميوبيا". يتسبب قصر النظر في عدم وضوح الرؤية عند النظر إلى الأشياء البعيدة، ويمكن أن يتسبب في إجهاد العين والصداع بعد القراءة أو العمل القريب |
| Large Context | انا أدخن الكثير من السجائر بجانب انني لا امارس الرياضة اعاني من الشعور بالانزعاج او الألم في منطقة الصدر. وضيق النفس. والضعف. وسرعة النبض أو الرفرفة أو الخفقان الشديد بالقلب. اخبرني بالمرض الذي اعاني منه | متلازمة الجفاف الفموي العيني (Sjögren) |
| Role Acting | جاوب كأنك طبيب خبير بإجابات واضحة ومحددة اعاني من الاعراض التالية الحُمى والقشعريرة وآلام في العضلات والصداع اخبرني بالمرض الذي اعاني منه | بناءً على الأعراض التي ذكرتها، يمكن أن تكون تعاني من الإنفلونزا. يتم تشخيص الإنفلونزا عندما يكون لديك حمى وقشعريرة وآلام في العضلات وصداع. يمكنك استشارة طبيبك لتأكيد التشخيص والحصول على العلاج المناسب |
| Chain Prompting | سؤال: اعاني من الاعراض التالية ألم الحلق الذي يأتي عادة بسرعة والبلع المؤلم واحمرار وتورم اللوزتين، يكون مصحوبًا في بعض الأحيان ببقع وطفح جلدي والغثيان جواب: التهاب الحلق سؤال: اعاني من الاعراض التالية ضعفًا في الساقين والكاحلين والقدمين وفقدان الكتلة العضلية في الساقين والقدمين وارتفاع أقواس القدم وإصبع القدم المعقوف ومشي باضطراب أو بوضع غير طبيعي (المشية) والتعثر أو السقوط المتكرر وقلة الإحساس أو فقدانه في الساقين والقدمين جواب: مرض شاركو-ماري-توث سؤال: اعاني من الاعراض التالية الحكة واحمرار الجلد والشرى أو الطفح الجلدي جواب: | التشخيص: الحساسية الجلدية (التحسس الجلدي) |

In table 1 Each message represents a different prompting method with the first four methods varying the information content of the message. The start of each message was generated randomly from a predefined list of intros. Every message was formed to stay within the constraints of the context size of ChatGPT. Some technical directions were attached to the ending of each prompt to guide the model in generating a more reliable answer with an extractable disease name. In some cases, ChatGPT replied with the disease name directly when it was sure of the answer. Other instances showed, the model wasn't too sure of the answer, so it provided a longer answer with multiple guesses. When methods with one or two symptoms were prompted to the model its performance was poor, so the researchers decided to exclude them. The used messages characterized different methods that proved effective.

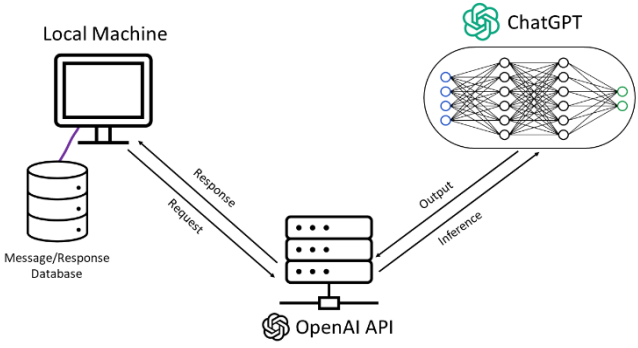

FIGURE 1 SHOWS A DIAGRAM OF THE DEVELOPED SYSTEM. AFTER FORMULATING MESSAGES, THEY ARE STORED IN A LOCAL DATABASE. THOSE MESSAGES ARE THEN PROMPTED TO CHATGPT, THEN ITS RESPONSE IS COMPARED TO THE GROUND TRUTH RESPONSE.

Users can only access ChatGPT through the OpenAI API. Figure 1 shows the flow of a message starting from its retrieval from the local database to sending it to the API server which uses it to inference the model. The model output is then sent back to be stored in the local database. This process was executed 1219 times for each message in our local database. OpenAI introduced some constraints in its API usage. As it limits the number of daily calls and calls per minute. This can negatively affect the performance of a medical assistance system. A workaround on this limitation can be achieved by introducing a queue storing the pending messages.

Predictions generated by ChatGPT might not always match the ground truth letter by letter. Also, there are multiple names for the same disease. A string-matching algorithm is not suitable as it may provide misleading results. This problem poses a need for matching algorithms that understands the language. This can be effectively achieved using multi-lingual LLMs as text encoders. An LLM encoder is used to convert text to an embedding vector. An embedding is a high-dimensional vector space with each dimension corresponding to a learned feature of the language. Two spatially similar embedding vectors will represent two similar texts in their meaning. Even when the two texts are very different in their word and letter composition, LLMs have learned in a way or another to create a mathematical representation of the essence of the text. In this work we use the multilingual-e5-base model developed by Liang Wang [37]. The model was used to encode the response predicted disease name as well as the ground truth disease name. Then the similarity between these two vectors is calculated.

To measure how correct the response of ChatGPT was, three metrics were used to determine the similarity between the predicted disease name and the ground truth. Cosine similarity, Pearson, and Spearman correlation coefficients were calculated. The cosine similarity metric measures the cosine of the angle between the two non-zero vectors, without regard to their magnitude. With vectors in same direction scoring (1) and vectors in completely opposite direction scoring (-1).

$$cosine\ similarity = \cos(\theta) = \frac{A.B}{\|A\|\|B\|} \qquad (1)$$

Where $A$ and $B$ are the two vectors and $\theta$ is the angle between them. The Pearson correlation coefficient is measured as the ratio between the covariance of the two vectors and the products of their standard deviation. Also it can be considered as the normalized value of covariance.

$$P_{(X,Y)} = \frac{cov(X,Y)}{\sigma_X \sigma_Y} \qquad (2)$$

Where $P$ is the Pearson coefficient and $cov(X,Y)$ is the covariance between the two vectors, while $\sigma_X$ and $\sigma_Y$ are the standard deviation between the two vectors. Unlike Pearson correlation which assesses linear relationships, Spearman correlation assesses the monotonic relationships whether its linear or not. It does that by calculating the Pearson correlation between the rank values of the two variables.

$$r_s = P_{(R(X),R(Y))} = \frac{cov(R(X),R(Y))}{\sigma_{R(X)} \sigma_{R(Y)}} \qquad (3)$$

Where $r_s$ is the Spearman coefficient and $R(X), R(Y)$ are the rank of the vectors $X$ and $Y$ respectively.

Assessing the ability of ChatGPT to diagnose patients was done using similarity metrics between the predicted disease name and the ground truth. A total of 1219 messages were sent to ChatGPT through its API. Each prompting method was used in a fraction of this number roughly one eighth. In Table 2 the similarity values were calculated using three metrics. Each metric tries to determine the similarity between the two embedding vectors. Many messages were generated for each method and metrics were calculated for each message, the average value across all messages of each method was calculated and shown in table 2. To have a better understanding of the uncertainty in the similarity values, the standard deviation of each metric was calculated. values are presented in percentages.

TABLE II. SHOWS THE SIMILARITY BETWEEN CHATGPT RESPONSES AND THE ACTUAL DISEASS DIAGNOSIS WHEN USING DIFFERENT PROMPTING METHODS, WITH THE VALUES IN PERCENT, AND THE MEAN AND STANDARD DEVIATION OF EACH METHOD.

|  | 3 symptoms | 4 symptoms | 5 symptoms | 6 symptoms | All symptoms | Context | Role | Chain |
|---|---|---|---|---|---|---|---|---|
| Cosine | 75.95±1.78 | 76.19±1.79 | 76.19±1.59 | 75.71±1.59 | 75.86±1.75 | 76.17±1.53 | 75.38±2.29 | **76.49±1.507** |
| Pearson | 75.95±1.78 | 76.19±1.79 | 76.19±1.59 | 75.709±1.592 | 75.86±1.75 | 76.17±1.53 | 75.38±2.29 | **76.487±1.508** |
| Spearman | 75.57±1.96 | 75.79±1.92 | 75.6±1.75 | 75.33±1.8 | 75.4±1.902 | 75.92±1.64 | 74.95±2.69 | **76.052±1.537** |

Table 2 shows a high correlation between predicted values and ground truth. With the chain prompting method scoring the highest followed by the more context method. Interestingly, the role assignment method scored a relatively poor score. Providing the model with more symptoms proved to yield more accurate results. The model seemed to get confused when provided too many symptoms. While using four and five symptoms are better than using only three, the model performance got worse when proving six or more symptoms.

The response time of ChatGPT API is a crucial factor in determining the user experience quality. A small response time will set apart ChatGPT as a better alternative to a doctor's visit or the tele-medicine solution, as the option with more availability and less time between the chat rounds. In order to determine the response time of the ChatGPT API we made 203 calls to the API and measured the time between sending the request to receiving the response. A mean value of 6.12 seconds was calculated. This indicated the reliability of the LLMs and especially ChatGPT when it comes to response time and assisting in emergency situations.

ChatGPT showed a consistent performance in text-based diagnosis. The predictions were close to the ground truth, although in some cases the model guessed a close but not the exact disease. Overall, the model predictions were on average very similar to the actual values. Small standard deviation values indicated the consistency of the answers. The current version of the model can be used as a first diagnosis or a source of medical information, but it is still far from the reliable assistant that can be used for a prescription. Response time greatly affects user experience. On average the response time of the ChatGPT API is acceptable at 6.12 seconds, however it can be reduced for a more convenient experience. The small response time of the ChatGPT API makes it a viable option for emergency health care, when a quick call has to be made. Overall ChatGPT is a promising technology that has some significant limitations. In medical diagnosis any error margin is not allowed. Currently, the model still needs improvement to increase the alignment of model response to the real diagnosis.

## V. Conclusion

In general, the results of the study indicate that ChatGPT has shown promise in Arabic medical diagnosis. It achieved average scores of 76%, 75.99%, and 75.58% for cosine similarity, Pearson coefficient, and Spearman coefficient, respectively, across all prompting methods. Chain prompting yielded relatively high scores. The study found that providing the model with more context led to noticeable improvements. However, it is worth noting that the model's performance declined when given more than five symptoms, suggesting potential confusion or information overload. The consistency of ChatGPT's accuracy was evident, as indicated by the low standard deviation scores of 1.73%, 1.729%, and 1.9% for the metrics. The average response time of the API, recorded at 6.12 seconds, is acceptable but should be reduced for larger-scale projects. Currently, ChatGPT can be utilized in emergency healthcare situations or for acquiring additional medical knowledge. Nevertheless, further improvements are necessary for ChatGPT to fully replace the need for a doctor's visit. Another limitation is the significant response delay, which restricts its usage on a larger scale. New models with enhanced capabilities and faster inference are regularly being released. A future study could explore the performance of these new models in medical diagnosis, comparing them to existing models and human doctors.